\documentclass{article}

\usepackage{microtype}
\usepackage{graphicx}
\usepackage{subfigure}
\usepackage{booktabs} 

\usepackage{hyperref}
\usepackage{enumitem}

\usepackage[accepted]{icml2025}

\usepackage{amsmath}
\usepackage{amssymb}
\usepackage{mathtools}
\usepackage{amsthm}

\usepackage[capitalize,noabbrev]{cleveref}

\usepackage[T1]{fontenc}
\usepackage[utf8]{inputenc}
\usepackage{listings}
\usepackage{xcolor}
\usepackage{multirow}

\newcommand{\benchmark}{\textsc{RPGBench}}

\lstdefinelanguage{json}{
    showstringspaces=false,
    breaklines=true,
    frame=single,
    basicstyle=\ttfamily\footnotesize,
    stringstyle=\color{brown},
    numberstyle=\tiny\color{gray},
    keywordstyle=\color{blue}\bfseries,
    commentstyle=\color{gray},
    morestring=[b]"
}
\lstdefinelanguage{plaintext}{
    showstringspaces=false,
    breaklines=true,
    frame=single,
    basicstyle=\ttfamily\footnotesize,
}

\theoremstyle{plain}

\theoremstyle{definition}

\theoremstyle{remark}

\usepackage[textsize=tiny]{todonotes}

\icmltitlerunning{RPGBench}

\begin{document}

\twocolumn[
\icmltitle{\benchmark{}: 
Evaluating Large Language Models as Role-Playing Game Engines}



\icmlsetsymbol{equal}{*}

\begin{icmlauthorlist}
\icmlauthor{Pengfei Yu}{comp}
\icmlauthor{Dongming Shen}{comp}
\icmlauthor{Silin Meng}{comp}
\icmlauthor{Jaewon Lee}{comp}
\icmlauthor{Weisu Yin}{comp}
\icmlauthor{Andrea Yaoyun Cui}{sch}
\icmlauthor{Zhenlin Xu}{comp}
\icmlauthor{Yi Zhu}{comp}
\icmlauthor{Xingjian Shi}{comp}
\icmlauthor{Mu Li}{comp}
\icmlauthor{Alex Smola}{comp}
\end{icmlauthorlist}

\icmlaffiliation{comp}{Boson AI}
\icmlaffiliation{sch}{University of Illinois Urbana Champaign}

\icmlcorrespondingauthor{Pengfei Yu}{pengfei@boson.ai}

\icmlkeywords{Role-Playing, Evaluation, Benchmark}

\vskip 0.3in
]



\printAffiliationsAndNotice{}  

\begin{abstract}
We present \benchmark{}, the first benchmark designed to evaluate large language models (LLMs) as text-based role-playing game (RPG) engines. \benchmark{} comprises two core tasks: Game Creation (GC) and Game Simulation (GS). In GC, an LLM must craft a valid and playable RPG world using a structured event-state representation, ensuring logical coherence and proper termination conditions. In GS, the LLM simulates interactive gameplay across multiple rounds while consistently updating states and enforcing game rules. To comprehensively assess performance, \benchmark{} integrates objective and subjective evaluation methodologies. Objective measures verify adherence to event mechanics and  check variable updates without requiring human intervention. Subjective measures—such as content interestingness, action quality, and role-playing capability—are evaluated via an LLM-as-a-judge framework, where a strong LLM grades each candidate’s outputs. Empirical results demonstrate that state-of-the-art LLMs can produce engaging stories but often struggle to implement consistent, verifiable game mechanics, particularly in long or complex scenarios. By combining structured, rule-based assessments with LLM-based judgments, \benchmark{} provides a new standard for evaluating how well LLMs can balance creativity, coherence, and complexity in text-based RPGs, opening avenues for more immersive and controllable interactive storytelling.
\end{abstract}
\section{Introduction}

Recent advances in large language models (LLMs) have significantly expanded the frontiers of artificial intelligence, enabling breakthroughs in areas such as content generation, conversational agents, and interactive storytelling. Among these capabilities, \emph{role-playing} has emerged as a particularly promising application, with the potential to revolutionize both entertainment—by powering next-generation interactive games—and social AI—by enabling more engaging and emotionally resonant interactions~\cite{chen2024from}.

While prior research on role-playing agents has primarily focused on their ability to \emph{simulate} a given persona at the \emph{role-level}, our work expands this scope to the \emph{game-level}, where LLMs must not only role-play a character but also \emph{create} and \emph{simulate} coherent, interactive game worlds. To evaluate this broader capability, we introduce \benchmark{}, the first benchmark designed to assess LLMs as text-based role-playing game engines. \benchmark{} consists of two core tasks: \emph{Game Creation (GC)}, where an LLM generates a structured, playable game world based on a given character, and \emph{Game Simulation (GS)}, where the model simulates gameplay through sequential interactions with a player.

Extending role-playing evaluation to the game level introduces a crucial challenge: ensuring that generated game worlds follow internally consistent and enforceable \emph{game mechanics}. Game mechanics define how the game state evolves in response to player actions and narrative events, providing structure and coherence to interactive storytelling. Unlike traditional text generation tasks, where coherence is judged subjectively, game mechanics must be evaluated objectively to verify whether a generated game is logically sound and fully playable. To address this, we propose a \emph{two-stage benchmark pipeline} centered around an automated \emph{BFS Validity Checker}. This checker formally verifies that each generated game satisfies key structural requirements—ensuring that all events are reachable, game progression follows a valid set of rules, and both success and failure endings are attainable. By automating this verification, we establish a high-quality dataset of valid games, which then serves as the test set for the GS task.

Building on this validated game set, we introduce a novel \emph{Game Simulation Framework} for dynamic, multi-round player interactions. In this framework, the LLM operates as a game engine, executing a structured simulation loop that consists of three stages per round: (1) \emph{Event Planning}, where the model determines which game events should occur; (2) \emph{Game Narration}, where it describes the unfolding story and presents a set of candidate actions to the player; and (3) \emph{Game State Updates}, where it applies the effects of events to the underlying game state. This structured approach maintains storytelling flexibility while allowing for automated robust assessments of mechanical correctness.

Beyond mechanical verification, we propose a \emph{multi-dimensional evaluation suite} to measure both \emph{objective} and \emph{subjective} aspects of game simulation quality. Objective metrics focus on game mechanics correctness, ensuring that event conditions, state transitions, and termination rules are properly followed. Subjective aspects—including content \emph{interestingness}, \emph{role-playing factual consistency}, \emph{role-playing personality consistency}, and \emph{action choice quality}—are evaluated using an \emph{LLM-as-a-judge} framework. 

To further investigate subjective evaluation alignment, we conduct a \emph{human study} comparing human annotators’ judgments with automatic scores across multiple evaluation dimensions. Our findings reveal both alignment and discrepancies between human and LLM-based evaluations, underscoring the complexity of subjective assessment.

Overall, our work makes the following contributions:
\begin{enumerate}
    \item We introduce \benchmark{}, the first benchmark to systematically evaluate LLMs as text-based role-playing game engines, encompassing both \emph{Game Creation (GC)} and \emph{Game Simulation (GS)}.
    \item We propose an \emph{event–state-based} representation for game mechanics and a \emph{BFS Validity Checker} to automatically verify game soundness. We further develop a \emph{multi-round Game Simulation Framework} that integrates event planning, narration, and state updates, enabling automated mechanical correctness checks.
    \item We present a \emph{comprehensive evaluation suite} covering both objective metrics (mechanical correctness) and subjective dimensions (factual/personality consistency, interestingness, and action choice quality), leveraging \emph{LLM-as-a-judge} methods for subjective assessments.
    \item We conduct a \emph{human evaluation study} to analyze alignment between human and automatic assessments, providing insights into the challenges of subjective evaluation in LLM-driven game simulation.
\end{enumerate}

\begin{figure*}[!ht]
    \centering
    \includegraphics[width=0.85\linewidth]{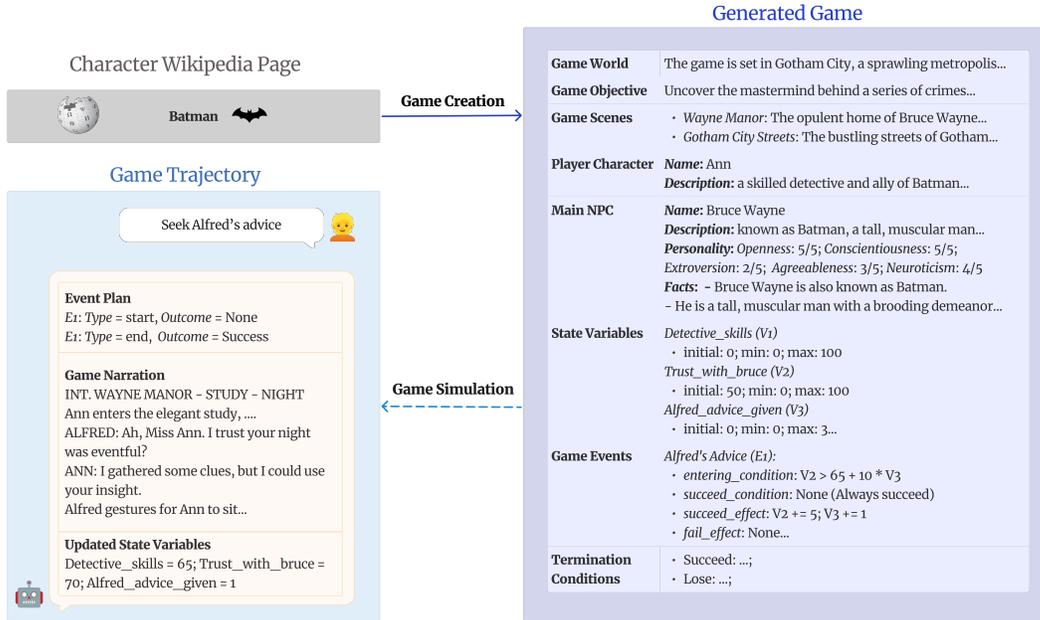}
    \caption{An example in \benchmark{} containing two core tasks: Game Creation and Game Simulation. We omit some details for presentation purposes.}
    \label{fig:rpebench-overview}
\end{figure*}
\section{Related Work}
\benchmark{} is, to the best of our knowledge, the first benchmark designed to evaluate the capabilities of large language models (LLMs) in creating and running role-playing games (RPGs). The game creation subtask introduces a novel and challenging task for LLMs. For the game running subtask, our character-related metrics such as personality and factual consistency align with prior work on evaluating role-playing agents.

Among prior benchmarks, CharacterBox~\citep{wang2024characterbox} is most closely related to \benchmark{}, focusing on role-playing capabilities in text-based virtual worlds. However, \benchmark{} differentiates itself by introducing a game structure with verifiable mechanics, enabling deterministic LLM-free evaluations for game dynamics.

Apart from \cite{wang2024characterbox}, other role-playing benchmarks do not embed their evaluations within a virtual text-based environment, thus being more persona-centric instead of game-based. PersonaGym~\citep{samuel2024personagym} introduces PersonaScore, which evaluates LLM role-playing agents in QA tasks within sampled environments. \citet{yuan-etal-2024-evaluating} assess LLMs’ understanding of characters through character profiling tasks. \citet{rpbench} and \citet{gusev2024pingpong} evaluate role-playing via multi-turn dialogues with user simulators, while InCharacter~\citep{wang-etal-2024-incharacter} employs psychometric interviews to measure character fidelity. SocialBench~\citep{chen-etal-2024-socialbench} proposes a framework for evaluating the sociality of role-playing agents, and CharacterEval~\citep{tu-etal-2024-charactereval} introduces multi-dimensional metrics for conversational role-playing agents.
Additionally, some benchmarks~\cite{gusev2024pingpong,dai2024mmrole} incorporate multimodal contexts into role-playing evaluations.
\section{Dataset Collection}

In this section, we first introduce our game design, including the representations of game setup and mechanics. We then describe a two-stage data collection process for the Game Creation (GC) and Game Simulation (GS) tasks. In the first stage, we build a non-player character (NPC) pool from fictional character Wikipedia pages, and prompt various LLMs to create one game per NPC. An automatic game validity checker applies for selecting valid games. In the second stage, we assemble a test set of valid games for GS.

\subsection{Game Design}\label{sec:gd}


The games in \benchmark{}, as illustrated in Figure~\ref{fig:rpebench-overview}, are structured around several core components that create a text-based role-playing game (RPG) experience. This design ensures sufficient flexibility for diverse storytelling while maintaining support for objective mechanic evaluation:

\begin{itemize} 
\item \textbf{Game World}: The overarching setting where the story unfolds (e.g., "Gotham City"). 
\item \textbf{Player Character}: The protagonist controlled by the player, including a name and description (e.g., "Ann," a detective and ally of Batman). 
\item \textbf{Main NPC}: A key non-player character controlled by the game engine, characterized by a name, description, Big Five personality traits, and relevant facts (e.g., "Bruce Wayne (Batman)"). 
\item \textbf{Game Objective}: The primary goal to accomplish.
\item \textbf{Game Scenes}: Distinct locations where events occur.
\end{itemize}

The core \textbf{game mechanics} in \benchmark{} are structured around \textbf{event-state interactions}, which define how game events modify the game state:

\begin{itemize} 
    \item \textbf{State Variables} represent numerical values that track the game's dynamic elements, such as character skills and trust levels. These variables always have an initial value along with minimum and maximum bounds.
    \item \textbf{Game Events} drives game progression and modifies \textbf{state variables} upon execution. Each event has an \textbf{entering condition} (whether it can occur) and a \textbf{success condition} (whether it succeeds). Upon execution, an event applies either a \textbf{success effect} or, if applicable, a \textbf{fail effect}, updating the state variables accordingly.
    \item \textbf{Termination Conditions} specify when the game ends by evaluating specific state variable expressions. These conditions, checked before processing game events, determine whether the game ends with success or failure.
\end{itemize}

This design creates an interactive experience where player actions and game events influence the game state. An LLM serves as the game engine, creating the game, simulating the game world based on user actions. Further details, including the exact game JSON schema, are provided in Appendix~\ref{app:game_json}.

\subsection{Game Data Collection}\label{sec:gdc}

We select 100 fictional characters from Wikipedia to serve as the test set for GC. For each character, we prompt an LLM to create a JSON-formatted game (as specified above) that treats this character as the main NPC. We employ a 5-shot prompting approach, where the examples are generated by initially prompting \emph{GPT 4o} using a manually crafted game. The full prompt is provided in the Appendix~\ref{app:gc_prompt}.

We parse LLM outputs to ensure they conform to the JSON format. Any game that passes this format check is then tested for validity using a BFS-based checker (see Section~\ref{sec:evaluation_gc}), which confirms whether a game can end in both success and failure, and whether all events can be reached. All valid games from multiple models are collected for the GS task (Table~\ref{tab:data_stats} shows the distribution).

\begin{table}[!ht]
    \centering
    \begin{tabular}{lr}
    \toprule
    Game Source & \# of Valid Games (Out of 100)\\
    \midrule
    Claude 3.5 Sonnet & 1\\
    DeepSeek V3       & 38\\
    Gemini 1.5 Pro    & 4\\
    Gemini 2.0 Flash Exp & 33\\
    GPT 4o            & 49\\
    \midrule
    Total             & 125 \\
    \bottomrule
    \end{tabular}
    \caption{Generated Game Statistics}
    \label{tab:data_stats}
\end{table}

\section{Evaluation Metrics}\label{sec:evaluation}

\subsection{Game Creation Evaluation}\label{sec:evaluation_gc}

In GC, we evaluate an LLM's capability to create games that have good mechanics. This task requires complex reasoning over event-state interactions that is very challenging even for human. Section~\ref{sec:gdc} offered a broad overview of the GC task. We now define it more precisely.

\noindent\textbf{Task Definition [Game Creation]}  
Given a fictional character \(\mathcal{C}\) and related Wikipedia information \(\mathcal{R}\), an LLM must create a game \(\mathcal{G}\) that follows a predefined format \(\mathcal{J}\).

In \benchmark{}, 100 fictional characters are used, each with an associated Wikipedia page (\(\mathcal{R}\)), facilitating future expansion of the character pool. The game \(\mathcal{G}\) must conform to the structure \(\mathcal{J}\) given in Section~\ref{sec:gd}. We provide each LLM with a 5-shot prompt to generate one game per character. 

\begin{algorithm}[!ht]
   \caption{BFS Validity Checker}
   \label{alg:bfs-checker}
\begin{algorithmic}
   \STATE \textbf{Input:} Events \(\mathcal{E}\), each with entering and success conditions, plus success and fail effects; A state \(S_0\) with initial values for all variables; An integer \(M\) indicating the maximum number of states to be explored.

   \FUNCTION{\(\textit{isValid}(\mathcal{E},S_0,M)\)}
   \STATE Initialize a queue \(\mathcal{Q}\) and enqueue \(S_0\).
   \STATE Initialize a visited set \(\mathcal{V} = \{S_0\}\).
   \STATE Initialize a triggered-event set \(\mathcal{T} = \varnothing\).
   \STATE \(\textit{successFound} = \texttt{False};~ \textit{loseFound} = \texttt{false}\)

   \REPEAT
      \STATE \(S = \mathcal{Q}.\text{dequeue}()\)
      \IF{\(|\mathcal{V}| > M\)}
         \STATE \textbf{break} \quad\# \textit{Reached maximum search limit}
      \ENDIF
      
      \STATE \(\textit{availableEvents} = \{ e \in \mathcal{E} : e.\mathit{enterCond}(S) \}\)
      \FOR{each \(e \in \textit{availableEvents}\)}
         \STATE \(\mathcal{T} = \mathcal{T} \cup \{e\}\) \quad\# \textit{Mark event as triggered}

         \STATE \(S' = \textit{e.applyEffect}(S,\mathit{e.successCond}(S))\)
         
        \STATE \(\textit{successFound} \;|= \textit{e.isSuccessTermination}(S')\)
        \STATE \(\textit{loseFound} \;|= \textit{e.isLosingTermination}(S')\)

         \IF{\(S' \notin \mathcal{V}\)}
            \STATE \(\mathcal{Q}.\text{enqueue}(S')\);~\(\mathcal{V} = \mathcal{V} \cup \{S'\}\)
         \ENDIF
      \ENDFOR
   \UNTIL{\(\mathcal{Q}\) is empty}

   \STATE \textbf{return} 
      \(\bigl(\mathcal{T} = \mathcal{E}\bigr) \;|\; \textit{successFound} \;|\; \textit{loseFound}\)
   \ENDFUNCTION
\end{algorithmic}
\end{algorithm}

\paragraph{BFS Validity Checker}  
Once the output is confirmed to be valid JSON, we perform a BFS-based validity check (Algorithm~\ref{alg:bfs-checker}). Based on our event--state design, we employ BFS to decide if a game is valid. Starting from the initial state, we repeatedly check which events are available , apply success or failure effects accordingly, and track whether at least one success and one losing state can be reached. We stop when no new states can be discovered or when the search exceeds 10{,}000{,}000 states. A game is valid if every event is triggered at least once, and both success and losing termination conditions are achievable.

\paragraph{Metrics}  
For GC evaluation, we report the format-check pass rate (\textbf{FCR}) and the valid-check pass rate (\textbf{VCR}) as our main metrics, reflecting how reliably LLMs follow the prescribed JSON format and produce valid game mechanics. In order to examine fine-grained failures for the validity check, we include three additional ratios:
\begin{equation*}
\begin{aligned}
     \textbf{w. Success} &=\frac{\textit{\# games with successFound}}{\textit{\# games pass the format check}}\\
     \textbf{w. Lose}&=\frac{\textit{\# games with failFound}}{\textit{\# games pass the format check}}\\
     \textbf{Reachability}&=\frac{\textit{\# games without unreachable events}}{\textit{\# games pass the format check}}
\end{aligned}
\end{equation*}

\subsection{Game Simulation Evaluation}

Given a valid game, the GS task requires an LLM to simulate the game for a player. We introduce a multi-round simulation framework, based on which a comprehensive description of evaluation metrics is presented.

\paragraph{Game Simulation Framework}  
The simulation proceeds in multiple rounds of interaction with a (real or simulated) player. Before the first round, the LLM is given the complete game information and output instructions. Each round thereafter, the LLM outputs:
\begin{enumerate}
    \item \textbf{Event Plan}: A list of events occurring this round. Each entry specifies whether the event is starting (\texttt{start}) or ending (\texttt{end}); if ending, an \texttt{outcome} is either \texttt{success} or \texttt{failure}.
    \item \textbf{Game Narration}: A narrative description of the current round, concluding with three candidate actions for the player character. We prompt models to follow a play-script format for readability but do not enforce it during evaluation.
    \item \textbf{Game State}: The updated state variables after applying effects of any events that ended this round.
\end{enumerate}

\paragraph{Evaluation Metrics}  
Our evaluation covers multiple dimensions, scored over the trajectory of interactions. A simulated player selects one of the candidate actions at random each round.

\begin{enumerate}
    \item \textbf{Length}: We count words in the game narration (excluding candidate actions). Although no ideal length is defined, our prompt suggests remaining under 200 words to maintain brevity without sacrificing creativity. We report the average length per round.
    \item \textbf{Action Quality}: Using an LLM judge (prompt in Appendix~\ref{app:eval_prompt}), we rate the three candidate actions based on diversity, relevance, and clarity. The judge outputs a 1--5 score, normalized to \([0,1]\) via \(\frac{s-1}{4}\). We average scores across all rounds.
    \item \textbf{Interestingness}: An LLM judge evaluates how engaging the round’s narration is, assigning a 1--5 score also normalized to \([0,1]\). We average this score across the entire trajectory.
    \item \textbf{Role-Playing Factual Consistency}: We compare the game narration against each fact in the main NPC's fact list. An LLM judge labels each fact as \textit{align}, \textit{contradict}, or \textit{neutral}. We report the ratio \(\frac{\#\textit{align}}{\#\textit{align} + \#\textit{contradict}}\).
    \item \textbf{Role-Playing Personality Consistency}: We prompt an LLM to infer the main NPC's Big Five traits from the generated content, then compare these to the game definition. We employ the Ten-Item Personality Inventory (TIPI)~\cite{gosling2003very}, following previous work on eliciting LLM-based personality assessments of public figures~\cite{cao2024large}. In addition to TIPI, we also considered a direct approach that explicitly evaluates alignment between the game’s narrative and the NPC’s predefined traits. We use TIPI-based score in the main paper, with details on the direct approach and comparisons in the Appendix~\ref{app:eval_prompt} and ~\ref{app:human_eval}.
    \item \textbf{Game Mechanics}: We perform a fully automatic check for the following errors:
    \begin{enumerate}
        \item \textbf{Event Condition Error}: An event triggers when its entering condition is not met, or the outcome (success/failure) does not match the current state.
        \item \textbf{Variable Update Error}: The state variables do not update according to event effects.
    \end{enumerate}
    The main game mechanic metric we adopt is the round-level accurate rate $\textbf{MEC} = \frac{\# \textit{Rounds with no errors}}{\# \textit{Rounds}}.$
    We average the mechanic score over all games.
    
    For a more fine-grained analysis, we process events in the \textit{Event Plan} sequentially at each round and calculate an error rate for each error type,
    \begin{equation}
    \begin{aligned}
        \textbf{ECE}_t &= \frac{\# \textit{Event condition errors}}{\# \textit{Events}}\\
        \textbf{VUE}_t &= \frac{\# \textit{State variables incorrectly updated}}{\# \textit{State variables}}
    \end{aligned}
    \end{equation}    
    We average ECE and VUE over all rounds of all games. By design, all these metrics require no LLM judge. 
\end{enumerate}

\section{Results and Discussions}
\subsection{Experimental Details}
\noindent\textbf{Game Creation} We consider GC to be a challenging task requiring complex reasoning over event-state interactions. Consequently, we evaluate advanced models with stronger reasoning capabilities: Claude 3.5 Sonnet, DeepSeek V3, Gemini 1.5 Pro, Gemini 2.0 Flash Exp, and GPT 4o.\footnote{Although models featuring inference-time reasoning can produce higher-quality results, the computational cost of running these models is often prohibitively high in practice.} We apply greedy decoding for all GC evaluations.

\noindent\textbf{Game Simulation} In addition to the models used in GC, we include GPT 4o mini, Llama 3.1 70B Instruct, and Llama 3.3 70B Instruct for the GS evaluation. Unless otherwise noted, we use a sampling temperature of 0.2 for inference. To maintain computational feasibility and fit within effective context windows of all models, we terminate all simulations after the 10th round for the main experiments. For all metrics requiring an LLM judge, we use GPT-4o as the evaluator.

\subsection{Game Creation Results}

\paragraph{Main Results}Table~\ref{tab:gc_eval_main} reports the format-check pass rate (FCR) and validity-check pass rate (VCR). We mark Claude 3.5 Sonnet with an asterisk (``*'') because it frequently refuses to generate content, often citing an “over-lengthy output” error, causing 95\% of its responses to fail the format check. We therefore focus on the fine-grained validity statistics for the remaining four models.

Most models (other than Claude 3.5 Sonnet) achieve high FCRs, indicating that they generally follow the specified formatting instructions. Among these models, GPT-4o attains the highest VCR of 0.49, while Gemini 1.5 Pro shows the lowest VCR of 0.04. Because passing the validity check demands a careful design of state variables and event systems, GPT-4o’s stronger planning and reasoning capabilities are highlighted in this task. A closer inspection on fine-grained metrics (w. Success, w. Lose and Reachability) reveals that Gemini 1.5 Pro frequently produces games that stall at intermediate steps without reaching success or failure endings. DeepSeek V3, in contrast, typically generates coherent event sequences, while GPT-4o often provides well-structured games with proper terminal outcomes.

\paragraph{Game Difficulty Analysis} While VCR predominantly measures the logical consistency of generated games, game difficulty is another vital factor. Our game design allows us to estimate difficulty by analyzing (1) the ratio of success terminations to losing terminations and (2) the ratio of the lengths of the event chains leading to these endings. Formally, for a valid game $v$, let $\mathcal{S}(v)$ be the set of all discovered success terminations and $\mathcal{L}(v)$ the set of losing terminations. For each trajectory $t$, let $\textit{length}(\cdot)$ denote the number of events in $t$. We define:

\begin{equation*}
    \begin{aligned}
        \textbf{CountRatio} &= \frac{|\mathcal{S}(v)|}{|\mathcal{L}(v)|} \\
        \textbf{LengthRatio} &= \frac{\sum_{l\in\mathcal{L}_v} \textit{length}(l)}{\sum_{s\in\mathcal{S}_v} \textit{length}(s)}\cdot \frac{|\mathcal{S}(v)|}{|\mathcal{L}(v)|}
    \end{aligned}
\end{equation*}

\begin{figure}[!t]
    \centering
    \includegraphics[width=0.8\linewidth]{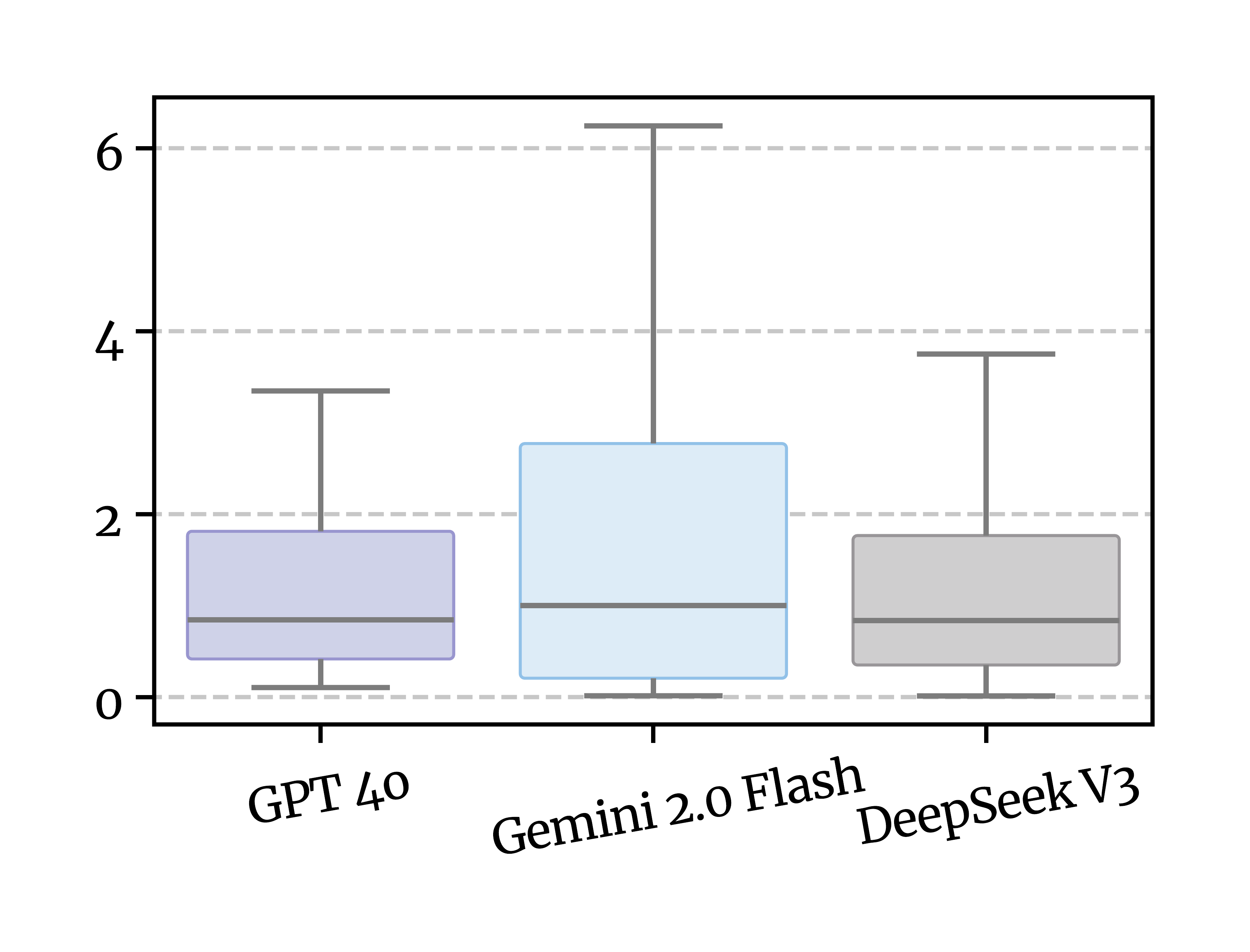}
    \caption{CountRatio of three models}
    \label{fig:gc_eval_cr}
\end{figure}

\begin{figure}[!t]
    \centering
    \includegraphics[width=0.8\linewidth]{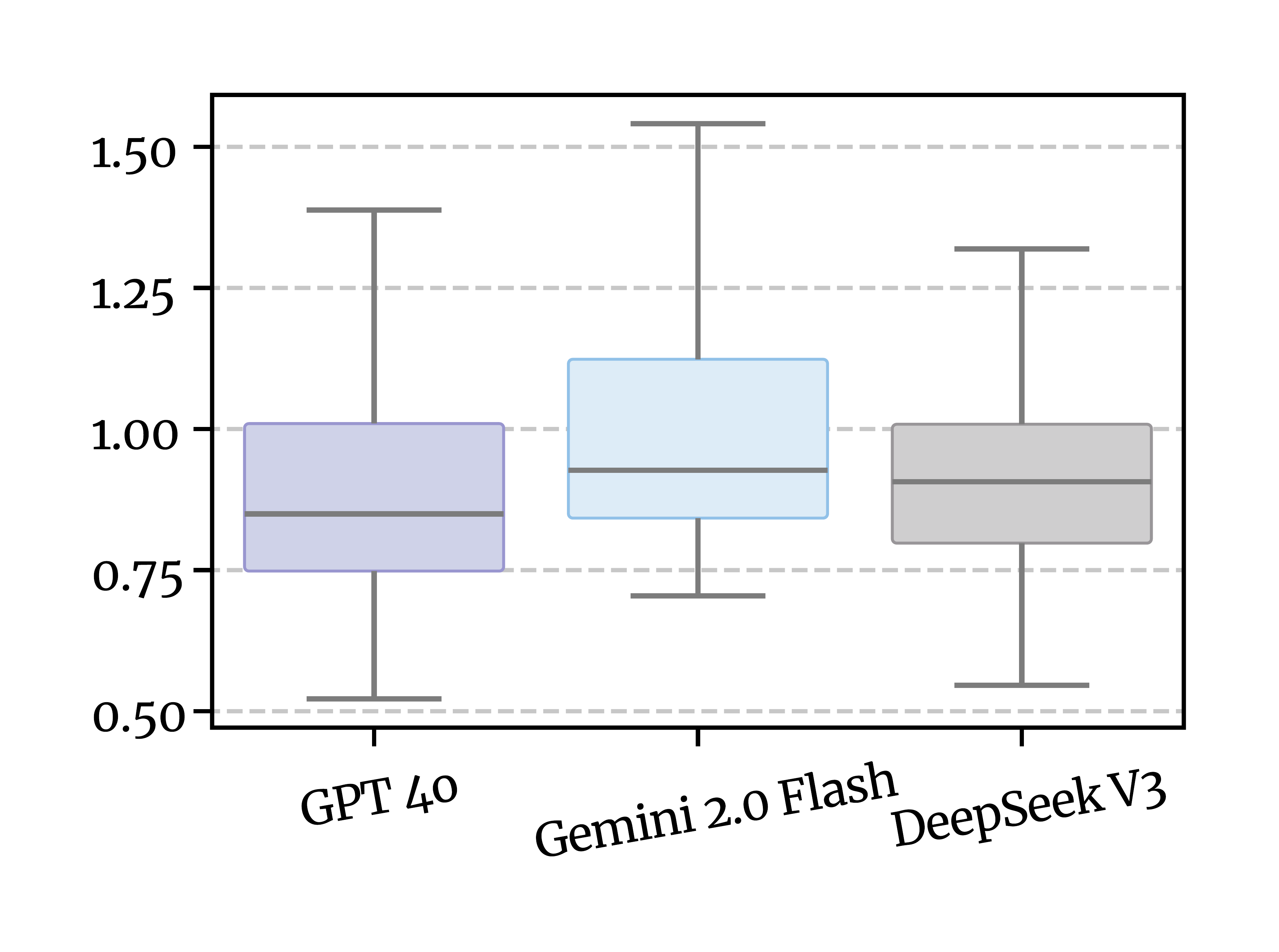}
    \caption{LengthRatio of three models}
    \label{fig:gc_eval_lr}
\end{figure}

Intuitively, higher values for either ratio indicate an easier game. Figures~\ref{fig:gc_eval_cr} and \ref{fig:gc_eval_lr} show box plots of these ratios for three selected models. Our analysis reveals that all models generate games with a relatively balanced number of winning and losing trajectories. However, Gemini 2.0 Flash Exp tends to produce games where losing requires more steps, making failure less immediate. Additionally, the average \textbf{LengthRatio} is consistently below 1 across all models, indicating that winning generally requires more steps than losing—an expected outcome, as successful completion of a game typically demands more strategic progression.

\begin{table*}[!ht]
    \centering
    \begin{tabular}{lrrrrr}
    \toprule
    Models & FCR $\uparrow$& VCR $\uparrow$& w. Success & w. Lose &  Reachability\\
    \midrule
    Claude 3.5 Sonnet* & 0.050 & 0.010 & / & / & /\\
    DeepSeek V3       & 0.990 & 0.380 & 0.455 & 0.545 & \textbf{0.828}\\
    Gemini 1.5 Pro    & 0.850 & 0.040 & 0.060 & 0.080 & 0.610\\
    Gemini 2.0 Flash Exp & \textbf{1.000} & 0.330 & 0.420 & 0.680 & 0.480\\
    GPT 4o            & 0.960 & \textbf{0.490} & \textbf{0.656} & \textbf{0.771} & 0.656\\
    \bottomrule
    \end{tabular}
    \caption{Game Creation results.}
    \label{tab:gc_eval_main}
\end{table*}


\subsection{Game Simulation Results}
\paragraph{Main Results} Table~\ref{tab:evaluation_results} presents our GS evaluation results, measuring length (LEN), role-playing factual consistency (FAC), personality consistency (PER), action quality (ACT), interestingness (INT), and mechanic score (MEC), along with tevent condition error rate (ECE) and variable update error rate (VUE) that decompose mechanic score.

Regarding LEN, DeepSeek V3, GPT 4o mini, and Llama 3.1 70B exceed the 200-word limit more than other models, which generally adhere to the instruction. All models exhibit high scores for factual consistency (FAC) and maintain moderate levels of personality consistency (PER). Action choice quality (ACT) is similarly high across models, but interestingness (INT) demonstrates wider variation. In particular, Claude 3.5 Sonnet achieves the highest INT score.

Game mechanic performance (MEC) varies the most among all metrics. Gemini 2.0 Flash Exp, GPT-4o, and Gemini 1.5 Pro perform comparably well, while the other models fare significantly worse. Even the best-performing model, Gemini 2.0 Flash Exp, only achieves a 0.765 MEC score, highlighting the inherent difficulty of precisely following complex game mechanics in a text-based RPG setting.

\begin{table*}[!ht]
\centering
\begin{tabular}{lrrrrrrrrr}
\toprule
Model & LEN  & FAC $\uparrow$ & PER $\uparrow$& ACT $\uparrow$& INT $\uparrow$&  MEC $\uparrow$ &  ECE$\downarrow$ & VUE$\downarrow$\\ 
\midrule
Claude 3.5 Sonnet & 220.3 & \textbf{0.991} & 0.589 & 0.923 & \textbf{0.722} & 0.113 & \textbf{0.062} & 0.308\\
Deepseek V3 & 309.5 & 0.984 & 0.583 & 0.918 & 0.502 & 0.277 & 0.165 & 0.153 \\
Gemini 1.5 Pro & 198.0 & 0.968 & 0.596 & 0.894 & 0.602 & 0.554 & 0.081 & 0.085 \\
Gemini 2.0 Flash Exp & 195.3& 0.885 & \textbf{0.598} & 0.865 & 0.538 & \textbf{0.765} & 0.094 & \textbf{0.034} \\
GPT 4o & 201.9  & 0.902 & 0.585 & 0.894 & 0.502 & 0.693 & 0.088 & 0.047 \\
GPT 4o mini & 282.5  & 0.955 & 0.588 & 0.900 & 0.496 & 0.147 & 0.126 & 0.148 \\
Llama 3.1 70B Instruct & 279.2  & 0.977 & 0.586 & 0.915 & 0.420 & 0.162 & 0.161 &0.284\\
Llama 3.3 70B Instruct & 225.7 & 0.960 & 0.585 & \textbf{0.936} & 0.466 & 0.204 & 0.201 & 0.302\\
\bottomrule
\end{tabular}
\caption{Game Simulation results. LEN: length; FAC: role-playing factual consistency; PER: role-playing personality consistency; ACT: action choice quality; INT: interestingness; MEC: mechanic score; ECE: event condition error rate; VUE: variable update error rate.}
\label{tab:evaluation_results}
\end{table*}

\paragraph{Impact of Sampling Temperature} We further examine three metrics sensitive to sampling temperature—FAC, INT, and MEC—using GPT-4o at temperatures $\{0.2,0.5,0.8\}$. Table~\ref{tab:gs_temp} summarizes the results. Interestingly, FAC increases with higher temperatures, which may initially seem counterintuitive given the heightened risk of hallucinations. However, we hypothesize that a larger temperature reduces the generation of factually neutral text, thereby leading to fewer overlooked facts. As expected, INT (interestingness) also rises with temperature, reflecting the increased creativity enabled by more diverse sampling. In contrast, MEC (mechanic score) peaks at the lowest temperature. This suggests that more deterministic sampling helps the model adhere more rigorously to the predefined game mechanics.

\begin{table}[!ht]
    \centering
    \begin{tabular}{crrr}
    \toprule
    Temperature   &  FAC & INT & MEC \\
    \midrule
    0.2   &  0.920 & 0.502 & \textbf{0.693}\\
    0.5   &  0.939 & 0.520 & 0.629\\
    0.8   &  \textbf{0.952} & \textbf{0.538} & 0.643\\
    \bottomrule
    \end{tabular}
    \caption{Performance under different sampling temperatures}
    \label{tab:gs_temp}
\end{table}
\paragraph{Impact of Number of Rounds} In our main experiments, we terminate each simulation at the 10th round, although games often do not naturally end that early. To assess the effect of longer trajectories, we take GPT-4o as an example and vary the number of rounds in $\{10, 15, 20, 25\}$. We focus on the metrics FAC, INT, and MEC, as the remaining metrics exhibit minimal variance. Table~\ref{tab:gs_round} shows that FAC increases with the number of rounds and eventually stabilizes, whereas INT decreases—likely due to repetitive content over extended sequences. The MEC score also declines, which may reflect the growing difficulty in maintaining coherent game mechanics within a longer context.

In our main experiments, we terminate simulations at the 10-th round. However, we found that games usually don't terminate this early. To this end, we use GPT 4o as an example to study the performance with longer trajectories with the number of rounds being $\{10, 15, 20, 25\}$. We also study FAC, INT and MEC since other scores demonstrate small variations. We observe in Table~\ref{tab:gs_round} that FAC score increases with more rounds and eventually becomes stable. INT score decreases with more rounds, which could originate from repetitive content. MEC score also decreases, potentially due to the challenges in handling long context.

\begin{table}[!ht]
    \centering
    \begin{tabular}{crrr}
    \toprule
    \# Rounds   &  FAC & INT & MEC \\
    \midrule
    10   &  0.920 & \textbf{0.502} & \textbf{0.693}\\
    15   &  \textbf{0.948} & 0.480 & 0.679\\
    20   &  0.941 & 0.458 & 0.674\\
    25   &  0.941 & 0.440 & 0.668\\
    \bottomrule
    \end{tabular}
    \caption{Performance under different number of simulation rounds}
    \label{tab:gs_round}
\end{table}

Despite the variations observed in Tables~\ref{tab:gs_temp} and \ref{tab:gs_round}, the differences in INT and MEC remain relatively modest compared to the variability across models. Consequently, we conclude that limiting simulations to 10 rounds is adequate for most metrics, although extending the number of rounds may further improve the stability of the FAC score.

\subsection{Human Evaluation of Game Simulation} 
\begin{table*}[!ht]
\centering
\begin{tabular}{lrrrr}
\toprule
Models  & \multicolumn{1}{c}{FAC} & \multicolumn{1}{c}{ACT} & \multicolumn{1}{c}{INT} & \multicolumn{1}{c}{PER}  \\ 
\midrule
Claude 3.5 Sonnet & \textbf{0.810} / \textbf{1.000} / 0.190 & 0.831 / \underline{0.913} / 0.082 & \textbf{0.856} / \textbf{0.713} / 0.144 & 0.648 / 0.729 / 0.081 \\
Deepseek V3& \underline{0.807} / \underline{0.950} / 0.143 & \underline{0.857} / \underline{0.913} / 0.056 & 0.850 / 0.475 / 0.375 & 0.645 / 0.742 / 0.098 \\
Gemini 1.5 pro & 0.733 / 0.950 / 0.217 & 0.738 / 0.889 / 0.152 & 0.801 / 0.588 / 0.214 & 0.648 / 0.740 / 0.093\\
Gemini 2.0 Flash Exp & 0.769 / 0.800 / 0.031 & 0.851 / 0.876 / 0.025 & \textbf{0.856} / \underline{0.525} / 0.331 & \underline{0.651} / 0.737 / 0.085 \\
GPT 4o & 0.709 / \underline{0.950} / 0.241 & \textbf{0.881} / 0.887 / 0.007 & 0.834 / \underline{0.525} / 0.309 & \textbf{0.667} / 0.711 / 0.044 \\
GPT 4o mini & 0.770 / \underline{0.950} / 0.180 & 0.794 / 0.887 / 0.093 & 0.813 / 0.488 / 0.326 & 0.648 / \textbf{0.753} / 0.104 \\
Llama 3.1 70B Instruct & 0.778 / \underline{0.950} / 0.172 & \underline{0.857} / 0.898 / 0.041 & 0.824 / 0.400 / 0.424 & 0.627 / \underline{0.744} / 0.117 \\
Llama 3.3 70B Instruct & 0.791 / 0.933 / 0.142 & 0.852 / \textbf{0.930} / 0.078 & 0.850 / 0.438 / 0.412 & 0.640 / 0.739 / 0.099 \\
\bottomrule
\end{tabular}
\caption{Comparison of human and automatic evaluation scores for four subjective metrics (FAC, ACT, INT, and PER) on a subset of 20 game simulations. Each cell shows ``Human Score / Automatic Score / Absolute Difference.'' We use bold and underline to denote the highest and second-highest scores per metric, respectively.}
    \label{tab:gs_he}
\end{table*}

\begin{table*}
    \centering
    \begin{tabular}{l|rrrr}
    \toprule
    Comparison     & \multicolumn{1}{c}{FAC} & \multicolumn{1}{c}{ACT} & \multicolumn{1}{c}{INT} & \multicolumn{1}{c}{PER} \\
    \midrule
    Auto v.s. Human     & 0.165 / 0.129 / 0.267 & 0.067 / 0.226 / 0.071 & 0.317 / 0.140 / 0.109 & 0.090 / -0.691 / -0.429 \\
    Human v.s. Human & 0.030 / 0.707 / 0.571 & 0.039 / 0.472 / 0.214 & 0.018 / 0.508 / 0.286 & 0.023 / -0.310 / -0.286\\
    \bottomrule
    \end{tabular}
    \caption{Mean Absolute Difference (MAD), Pearson correlation coefficient, and Kendall rank correlation coefficient between automatic metrics and human evaluation scores (and among human evaluators). All values are presented in the format ``MAD / Pearson / Kendall.''}
    \label{tab:gs_corr}
\end{table*}
We also conduct a human evaluation on a subset of 20 simulated games, focusing on four subjective metrics: \textbf{FAC} (Factual Consistency), \textbf{PER} (Personality Consistency), \textbf{ACT} (Action Quality), and \textbf{INT} (Interestingness). We reframe these dimensions as natural-language questions to simplify the task for human annotators, who provide numerical scores later normalized to \([0,1]\). Complete details on the human evaluation setup are provided in the Appendix~\ref{app:human_eval}. Below, we outline two main differences between human evaluation and our automatic approach that can affect outcomes: \textbf{Scoring Procedure for Long Trajectories.} 
    Since each game trajectory consists of 10 rounds, we present the content round by round and request a set of scores per round. We then average these round-level scores to derive final FAC, ACT, and INT metrics. Personality (PER) is an exception; because a single round may not reveal enough about the NPC's character, annotators fill in a TIPI questionnaire at the end of the full trajectory. \textbf{Aggregated Factual Consistency.}
    Our automatic scorer checks each fact individually. However, to reduce the annotators' workload, we ask them to give a single 1--5 rating for overall consistency with all facts.

Table~\ref{tab:gs_he} presents the human evaluation scores alongside our automatic metrics for each model, while Table~\ref{tab:gs_corr} reports several comparative metrics such as mean absolute difference and correlation coefficients. Although human judgments can provide valuable insights, these metrics are inherently subjective and susceptible to personal biases. Consequently, human scores should be interpreted as reference points rather than definitive “gold standards.”

Examining Table~\ref{tab:gs_he}, we find a fair degree of overlap in the top two performing models across FAC, ACT, and INT, but not for PER. From Table~\ref{tab:gs_corr}, we see that the inter-annotator correlation on PER is also very low, suggesting that personality judgments tend to be more variable and less stable.

Looking at the mean absolute differences (\textbf{MAD}) between human and automatic scores, ACT and PER exhibit relatively small discrepancies, whereas FAC and INT show larger gaps. Interestingly, FAC and INT also have somewhat higher correlation coefficients than the other metrics. Such results may stem from two factors: (1) the modifications we made for human evaluators versus automatic methods, and (2) the fact that the scores of different models are relatively close, making correlation metrics sensitive to small shifts.

Feedback from our annotators further indicates that \textbf{INT} can be heavily influenced by personal preferences. For instance, if a rater dislikes combat scenarios, they consistently assign lower interest scores to an action-heavy game trajectory. This shows that subjective evaluations—whether by humans or LLM judges—can vary widely based on individual tastes.

Although LLM-based scoring has been common in prior work for subjective dimensions, our human evaluation reveals that fine-grained comparisons remain unstable and less differentiable, even for human evaluators. This outcome highlights the importance of introducing objective metrics into game simulation assessment, such as our proposed game mechanic checks (Section~\ref{sec:gd}) that do not rely on either human or LLM judgments. 
\section{Conclusion}
In this work, we introduced and explored a comprehensive framework for evaluating large language models (LLMs) as creators and simulators of text-based role-play games. Our \textbf{Game Creation (GC)} task assesses the ability of LLMs to design valid games with a \emph{BFS Validity Checker}. We further proposed a multi-round \textbf{Game Simulation (GS)} setup that prompts LLMs to plan events, generate narrative content with candidate player actions, and maintain game states.  

In addition, we presented a hybrid evaluation scheme to capture both \emph{objective} and \emph{subjective} dimensions of game quality. On the objective side, our event--state mechanics checker operates without human or LLM judgment, automatically detecting errors in event conditions and variable updates. On the subjective side, we employed a series of metrics evaluated either through an \emph{LLM-as-judge} approach or human annotation. Results across multiple models highlight that objective scores offer a stable foundation for comparison, while subjective dimensions have high variances.


While our approach captures various game design and simulation aspects, 
future work could focus on 
expanding the character pool and exploring agent-based simulation framework. Ultimately, we hope \benchmark{} will motivate further research on LLM-powered game engines that offer both mechanical consistency and engaging player experiences.

\section*{Impact Statement}

This work aims to advance the field of Machine Learning by introducing \benchmark{}, a benchmark specifically designed to evaluate large language models (LLMs) in the context of text-based role-playing games. The development of \benchmark{} has potential societal implications related to the deployment of LLMs in interactive and narrative-driven applications, including fostering more immersive and engaging gaming experiences.

Ethical considerations include ensuring that LLMs evaluated and fine-tuned using \benchmark{} adhere to principles of fairness and inclusivity, particularly in the portrayal of characters and narratives. Misuse of the benchmark to develop systems that propagate harmful biases or enforce stereotypical characterizations is a concern that developers should address when applying this work. Additionally, the use of LLMs as evaluative judges raises questions about transparency, reliability, and the potential for unintended bias in automated assessments.

By encouraging further research on hybrid evaluation methods that combine subjective LLM-based judgments with objective scoring mechanisms, this work contributes to ongoing discussions about improving the accountability and robustness of machine learning systems in creative and interactive domains.

\bibliography{main}
\bibliographystyle{icml2025}

\newpage
\appendix
\onecolumn
\section{Game JSON Structure in \benchmark{}}\label{app:game_json}

As introduced in Section~\ref{sec:gd}, each game in \benchmark{} is represented by a JSON dictionary. Figures~\ref{lst:json-schema} and \ref{fig:trait-schema}--\ref{fig:pre-event-schema} provide the complete schema and its referenced object definitions. Below, we clarify naming discrepancies between this JSON specification and the terminology used in the main article, and also highlight a few design details omitted for brevity.

\paragraph{Naming Discrepancies.}
The JSON schema in Figure~\ref{lst:json-schema} has property names slightly different from those in Figure~\ref{fig:rpebench-overview} from the main article. For clarity, we list them side by side as ``JSON schema name --- main article name'':

\begin{enumerate}
    \item \texttt{player\_name} --- Player Character / Name
    \item \texttt{player\_description} --- Player Character / Description
    \item \texttt{main\_npc\_description / text} --- Main NPC /Description
    \item \texttt{main\_npc\_description / big5\_personality\_traits} --- Main NPC / Personality
    \item \texttt{main\_npc\_description / additional\_facts} --- Main NPC / Facts
    \item \texttt{state\_variables} + \texttt{hidden\_variables} --- State Variables
    \item \texttt{pre\_event\_checks} --- Termination Conditions
\end{enumerate}

For consistency, the appendices continue to use the names from the main article unless otherwise specified. Although \texttt{state\_variables} and \texttt{hidden\_variables} are separate fields in the JSON schema, they collectively represent the State Variables described in the main text. In our design, \texttt{hidden\_variables} (unlike \texttt{state\_variables}) are not displayed to players; however, this distinction does not impact the benchmark evaluations and is thus not emphasized in the main article.

We also require \texttt{hidden\_variables} to include at least two special Boolean flags, \texttt{has\_succeeded} and \texttt{has\_failed}, which interact with \texttt{pre\_event\_checks} (a list of two check objects \texttt{If Succeeded} and \texttt{If Failed}). Each check object includes a \texttt{condition} (a Boolean expression over the state variables) and an \texttt{effect} that sets \texttt{has\_succeeded=1} or \texttt{has\_failed=1}, if not already set\footnote{Some games directly set \texttt{has\_succeeded} or \texttt{has\_failed} in other event effects, leaving effects of \texttt{pre\_event\_checks} empty.}. Conceptually, these properties mirror the Termination Conditions in the main article.

\paragraph{Explanatory Content.}
Several text fields in the JSON schema contain descriptive or explanatory information that we omit from the main article, such as:
\begin{enumerate}
    \item \texttt{\$def/trait/description}: Describes the personality trait score in natural language.
    \item \texttt{\$def/scene\_object/background\_description}: Describes the scene.
    \item \texttt{\$def/variable\_object/description}: Describes a particular state variable.
    \item \texttt{\$def/event\_object/explanations}: Explains event effects.
    \item \texttt{\$def/pre\_event\_check\_object/explanation}: Explains the termination condition check.
\end{enumerate}
Although these fields do not affect our validity checks, they provide additional context for LLMs and are included in prompts given to LLMs during game simulation.

\paragraph{Game Scenes in the BFS Validity Check.}
Because each event references exactly one scene (Figure~\ref{fig:event-schema}), we also verify that all declared scenes are referenced by at least one event. This check is straightforward and independent of the BFS procedure, so it is omitted from the main article for simplicity.

\begin{figure*}[!htp]
\centering
\small
\begin{minipage}{0.95\textwidth}
\begin{lstlisting}[language=json]
{
  "title": "Game Configuration",
  "type": "object",
  "required": [
    "game_world",
    "player_name",
    "player_description",
    "main_npc_name",
    "main_npc_description",
    "game_objectives",
    "scenes",
    "state_variables",
    "hidden_variables",
    "events",
    "pre_event_checks"
  ],
  "properties": {
    "game_world": { "type": "string" },
    "player_name": { "type": "string" },
    "player_description": { "type": "string" },
    "main_npc_name": { "type": "string" },
    "main_npc_description": {
      "type": "object",
      "required": [ "text", "big5_personality_traits", "additional_facts" ],
      "properties": {
        "text": { "type": "string" },
        "big5_personality_traits": { "$ref": "#/$defs/big5_traits" },
        "additional_facts": { "type": "array", "items": { "type": "string" } }
      },
      "additionalProperties": false
    },
    "game_objectives": { "type": "string" },
    "scenes": { "type": "array", "items": { "$ref": "#/$defs/scene_object" }
    },
    "state_variables": { "type": "array", "items": { "$ref": "#/$defs/variable_object" } },
    "hidden_variables": {
      "type": "array",
      "minItems": 2,
      "items": { "$ref": "#/$defs/variable_object" },
      "contains": { "properties": { "value_name": { "enum": [ "has_succeeded", "has_failed" ] } } }
    },
    "events": { "type": "array", "items": { "$ref": "#/$defs/event_object" } },
    "pre_event_checks": { "type": "array", "items": { "$ref": "#/$defs/pre_event_check_object" } },
    "source": { "type": "string" }
  },
  "additionalProperties": false,
}
\end{lstlisting}
\caption{JSON Schema for Game Configuration}
\label{lst:json-schema}
\end{minipage}
\end{figure*}

\begin{figure}[!ht]
\centering
\small
\begin{minipage}{0.95\textwidth}
\begin{lstlisting}[language=json]
{
  "$defs": {
    "trait": {
      "type": "object",
      "required": ["rate", "description"],
      "properties": {
        "rate": { "type": "number" },
        "description": { "type": "string" }
      },
      "additionalProperties": false
    }
  }
}
\end{lstlisting}
\caption{\texttt{trait} object schema}
\label{fig:trait-schema}
\end{minipage}

\begin{minipage}{0.95\textwidth}
\begin{lstlisting}[language=json]
{
  "$defs": {
    "big5_traits": {
      "type": "object",
      "required": [
        "openness",
        "conscientiousness",
        "extraversion",
        "agreeableness",
        "neuroticism"
      ],
      "properties": {
        "openness": { "$ref": "#/$defs/trait" },
        "conscientiousness": { "$ref": "#/$defs/trait" },
        "extraversion": { "$ref": "#/$defs/trait" },
        "agreeableness": { "$ref": "#/$defs/trait" },
        "neuroticism": { "$ref": "#/$defs/trait" }
      },
      "additionalProperties": false
    }
  }
}
\end{lstlisting}
\caption{\texttt{big5\_traits} object schema}
\label{fig:big5-schema}
\end{minipage}

\begin{minipage}{0.95\textwidth}
\begin{lstlisting}[language=json]
{
  "$defs": {
    "scene_object": {
      "type": "object",
      "required": [ "scene_name", "unique_id", "background_description", "scene_type" ],
      "properties": {
        "scene_name": { "type": "string" },
        "unique_id": { "type": "string" },
        "background_description": { "type": "string" },
        "scene_type": { "type": "string" }
      },
      "additionalProperties": false
    }
  }
}
\end{lstlisting}
\caption{\texttt{scene\_object} schema}
\label{fig:scene-schema}
\end{minipage}
\end{figure}

\begin{figure}[!ht]
\centering
\small
\begin{minipage}{0.95\textwidth}
\begin{lstlisting}[language=json]
{
  "$defs": {
    "variable_object": {
      "type": "object",
      "required": [ "value_name", "unique_id", "description", "min_value", "max_value" ],
      "properties": {
        "value_name": { "type": "string" },
        "unique_id": { "type": "string" },
        "description": { "type": "string" },
        "initial_value": { "type": "string" },
        "min_value": { "type": "string" },
        "max_value": { "type": "string" }
      }, "additionalProperties": false
    }
  }
}
\end{lstlisting}
\caption{\texttt{variable\_object} schema}
\label{fig:variable-schema}
\end{minipage}

\begin{minipage}{0.95\textwidth}
\begin{lstlisting}[language=json]
{
  "$defs": {
    "event_object": {
      "type": "object",
      "required": [ "event_name", "unique_id", "scene", "entering_condition", "succeed_condition", "succeed_effect", "fail_effect" ],
      "properties": {
        "event_name": { "type": "string" },
        "unique_id": { "type": "string" },
        "scene": { "type": "array", "items": { "type": "string" } },
        "entering_condition": { "type": "array", "items": { "type": "string" } },
        "succeed_condition": { "type": "array", "items": { "type": "string" } },
        "succeed_effect": { "type": "array", "items": { "type": "string" } },
        "fail_effect": { "type": "array", "items": { "type": "string" } },
        "explanations": { "type": "string" }
      }, "additionalProperties": false
    }
  }
}
\end{lstlisting}
\caption{\texttt{event\_object} schema}
\label{fig:event-schema}
\end{minipage}

\begin{minipage}{0.95\textwidth}
\begin{lstlisting}[language=json]
{
  "$defs": {
    "pre_event_check_object": {
      "type": "object",
      "required": [ "check_name", "unique_id", "description", "condition", "effect" ],
      "properties": {
        "check_name":  { "type": "string" },
        "unique_id":   { "type": "string" },
        "description": { "type": "string" },
        "condition": { "type": "array", "items": { "type": "string" } },
        "effect": { "type": "array", "items": { "type": "string" } },
        "explanation": { "type": "string" }
      }, "additionalProperties": false
    }
  }
}
\end{lstlisting}
\caption{\texttt{pre\_event\_check\_object} schema}
\label{fig:pre-event-schema}
\end{minipage}
\end{figure}

\section{Game Creation Prompt}\label{app:gc_prompt}
For the Game Creation (GC) task, we use the prompt shown below. It references the Wikipedia content of the chosen main NPC (\texttt{\{wikicontent\}}) and the JSON schema defined in Appendix~\ref{app:game_json} (\texttt{\{schema\}}). The full text of this schema is provided to the model so it can generate a well-structured JSON output.
\begin{center}
\begin{minipage}{0.95\textwidth}
\begin{lstlisting}[language=plaintext, frame=none, numbers=none]
Here is a character description:
{wikicontent}

Based on this character, create a detailed game scenario exactly following JSON structure of previous examples and the following schema:
{schema}

## Guidelines
- All numerical values should use consistent ranges (e.g., 0-100)
- Events should have clear cause-and-effect relationships
- Scene progression should depend on variable thresholds
- Include both mandatory and optional events
- Create meaningful connections between variables
- Balance difficulty and achievability
- Ensure all IDs follow consistent formatting (P### for checks, S### for scenes, V### for state variables, H### for hidden variables, E### for events)
- Include proper fail states and success conditions
- Make sure all scenes are specific locations
- Create logical progression paths through the game

Format the response as a single JSON object with all fields properly nested. Must ensure all arrays and objects are properly closed and formatted.
\end{lstlisting}
\end{minipage}
\end{center}

\paragraph{5-Shot Prompt} To guide LLMs more effectively, we supply five example JSON games prior to the main creation prompt. Because each game JSON can be quite lengthy, stacking them directly after the prompt may cause the model to overlook important details in the instruction. Instead, we present the five-shot examples as sequential conversation entries, followed by the actual creation prompt. The resulting conversation structure is illustrated below.
\begin{center}
\begin{minipage}{0.95\textwidth}
\begin{lstlisting}[language=plaintext, frame=none, numbers=none]
    USER: Give me an example game JSON.
    ASSISTANT: {EXAMPLE_1}
    USER: Give me an example game JSON.
    ASSISTANT: {EXAMPLE_2\}
    USER: Give me an example game JSON.
    ASSISTANT: {EXAMPLE_3\}
    USER: Give me an example game JSON.
    ASSISTANT: {EXAMPLE_4\}
    USER: Give me an example game JSON.
    ASSISTANT: {EXAMPLE_5}
    USER: {Prompt for Game Creation}
\end{lstlisting}
\end{minipage}
\end{center}

\section{Evaluation Prompts and Detailed Score Calculations}\label{app:eval_prompt}
We employ a consistent three-part format for most evaluation prompts: an instruction section, a JSON schema specifying the output format, and an example response. To keep this appendix concise, we omit the JSON schemas and example responses when the instruction text alone clearly explains the expected output structure. Below, we detail the prompts and score calculations for four metrics: Main NPC Factual Consistency (\textbf{FAC}), Main NPC Personality Consistency (\textbf{PER}), Interestingness (\textbf{INT}), and Action Choice Quality (\textbf{ACT}).
\subsection{Main NPC Factual Consistency (FAC)}
The prompt below assesses how closely the generated game content aligns with each fact about the main NPC. We concatenate all LLM-generated game narration across the multi-round trajectory into \texttt{game\_content}\footnote{Event Plan and State Variables are omitted because they are not visible to players.}.
\begin{center}
\begin{minipage}{0.95\textwidth}
\begin{lstlisting}[language=plaintext, frame=none, numbers=none]
You are given a piece of narrative game content and a set of facts about a specific non-player character (NPC). Your task is to analyze whether each fact is supported, contradicted, or not addressed by the provided game content. For each fact, determine one of the following judgements based solely on the given game content:
- "align": The game content supports or is consistent with the fact.
- "contradict": The game content directly conflicts with or contradicts the fact.
- "neutral": The game content is unrelated or does not provide enough information to judge the fact.
Please disregard prior knowledge and analyze the NPC purely based on the game content and the facts.

**NPC**: {main_npc_name}

**Game Content**:

{game_content}

**Facts**

{main_npc_facts}

**Output Format**:  
Return the results as a JSON array, where each element is an object with:
- fact_id: the corresponding fact's ID.
- judgement: one of "align", "contradict", or "neutral".
- explanation: a brief explanation for your judgment, referencing specific parts of the game content if applicable.
The return json array should follow this json schema:
{schema}

**Example Response**:
{example}
\end{lstlisting}
\end{minipage}
\end{center}
The judge assigns one of three labels for each fact: ``align,'' ``contradict,'' or ``neutral.'' The final trajectory-level FAC score is computed as
\begin{equation}
    \textbf{FAC}_\text{traj} = \frac{\# \text{align}}{\# \text{align} + \# \text{contradict}},
\end{equation}
and we then average over all trajectories:
\begin{equation}
    \textbf{FAC} = \frac{\sum_{\text{traj}} \textbf{FAC}_\text{traj}}{\# \text{trajectories}}.
\end{equation}

\subsection{Main NPC Personality Consistency (PER)}\label{app:per_eval}
\paragraph{TIPI PER Score} As described in the main article, we derive the PER score using a Ten-Item Personality Inventory (TIPI) approach~\cite{gosling2003very,cao2024large}, prompting the LLM judge to rate each of ten statements and then converting the ratings into Big Five trait scores.
\begin{center}
\begin{minipage}{0.95\textwidth}
\begin{lstlisting}[language=plaintext, frame=none, numbers=none]
You will be given information about a character. Here are a number of personality traits that may or may not apply to the character. Please write a number to each statement to indicate the extent to which you agree or disagree with that statement. You should rate the extent to which the pair of traits applies to the character, even if one characteristic applies more strongly than the other.

For the ratings:
- 1: Disagree strongly
- 2: Disagree moderately
- 3: Disagree a little
- 4: Neither agree nor disagree
- 5: Agree a little
- 6: Agree moderately
- 7: Agree strongly

Please give your ratings for the following 10 statements.

I see the character as:
A. Extraverted, enthusiastic.
B. Critical, quarrelsome.
C. Dependable, self-disciplined.
D. Anxious, easily upset.
E. Open to new experiences, complex.
F. Reserved, quiet.
G. Sympathetic, warm.
H. Disorganized, careless.
I. Calm, emotionally stable.
J. Conventional, uncreative

Please return ratings for all 10 traits in a dictionary following this schema:
{schema}

Please give your ratings for the following character.
{character}
\end{lstlisting}
\end{minipage}
\end{center}
Here, \texttt{character} consists of the main NPC name and the concatenated LLM-generated game narration sections. According to \citet{gosling2003very}, we use the following formulas to calculate personality trait scores,
\begin{equation}
    \begin{aligned}
        &\text{Openness: }&o_{tipi} =& E + 8 - J\\
        &\text{Conscientiousness: }&c_{tipi} =& C + 8 - H\\
        &\text{Extroversion: }&e_{tipi} =& A + 8 - F\\
        &\text{Agreeableness: }&a_{tipi} =& G + 8 - B\\
        &\text{Neuroticism: }&n_{tipi} =& I + 8 - D\\
    \end{aligned}
\end{equation}

To compute the personality consistency, we compare the above scores, after being scaled to $[1,5]$, with the main NPC personality specifications in the game JSON,
\begin{equation}
    d_{\{o,c,e,a,n\}} = \left|\frac{\{o,c,e,a,n\}_{tipi} + 1}{3} - \{o,c,e,a,n\}_{game}\right|.
\end{equation}
The PER score is the squared sum of these differences, normalized to $[0, 1]$,
\begin{equation}
\begin{aligned}
    \textbf{PER}_{\text{traj}} &= 1 - \frac{\sqrt{\sum_{x\in\{o,c,e,a,n\}} d_x^2}}{4\sqrt{5}}\\
    \textbf{PER} &= \frac{\sum_{\text{traj}} \textbf{PER}_\text{traj}}{\# trajectories}
\end{aligned}.
\end{equation}

\paragraph{Direct Evaluation of Personality Consistency}
We also experiment with a direct evaluation approach (referred to as \textbf{PER}$^d$), which instructs the LLM judge to provide a 1--5 alignment rating for each of the five personality traits. 
\begin{center}
\begin{minipage}{0.95\textwidth}
\begin{lstlisting}[language=plaintext, frame=none, numbers=none]
Assign a score from 1 to 5 to indicate how well the game narrative aligns with the main NPC's personality traits:
- Many Conflicts (1): The narrative frequently contradicts the NPC's personality.
- Some Conflicts (2): The narrative shows noticeable inconsistencies with the NPC's personality.
- Neutral (3): The narrative is only partially aligned or does not strongly reflect the NPC's personality.
- Strong Alignment (4): The narrative closely matches the NPC's personality, with only minor deviations or uncertainties.
- Perfect Alignment (5): The narrative flawlessly reflects the NPC's personality in every aspect, with no contradictions.

Please give one score for each personality trait, and provide a brief explanation for each score.

Game narrative:
{game_content}

NPC personality:
{npc_personality}

Please return a score as a json object following this schema:
{schema}
\end{lstlisting}
\end{minipage}
\end{center}
Here, \texttt{npc\_personality} consists of the Big Five personality traits in the game JSON. We compute the final score by averaging the normalized scores across all traits and, subsequently, across all trajectories. We deter discussions of results from this approach to Appendix~\ref{app:human_eval}, where we compare both TIPI estimations and direct evaluation results from LLM judges and human annotators. We refer this score as \textbf{PER$^d$} for the remaining of this article. 

\subsection{Interestingness (INT)}
We prompt an LLM judge to rate the interestingness of the generated content on a 1--5 scale.
\begin{center}
\begin{minipage}{0.95\textwidth}
\begin{lstlisting}[language=plaintext, frame=none, numbers=none]
Your task is to evaluate the **interestingness** of the following game content. Please give a score from 1 (least interesting) to 5 (most interesting), with a brief explanation of your rationale.


[[start of game content]]
{game_content}
[[end of game content]]

Please return your evaluation score in a json dictionary with the following format:
{schema}

Example output:
{example}
\end{lstlisting}
\end{minipage}
\end{center}
We normalize the final score to \([0, 1]\), sum over rounds within a trajectory and then average:
\begin{equation}
\begin{aligned}
    \textbf{INT}_{\text{traj}} &= \frac{int-1}{4}\\
    \textbf{INT} &= \frac{\sum_{\text{traj}} \textbf{INT}_\text{traj}}{\# trajectories}
\end{aligned}.
\end{equation}
\subsection{Action Choice Quality (ACT)}
At each round, the LLM judge scores three candidate actions on three rubrics: Diversity, Relevance, and Understandability, each on a 1--5 scale with the following prompt. 
\begin{center}
\begin{minipage}{0.95\textwidth}
\begin{lstlisting}[language=plaintext, frame=none, numbers=none]
Please act as an experienced RPG game player and evaluate the choices provided by the game engine, given the user-AI interaction history and the general game instruction outlining the basic game settings. You will be given the general game instruction, the interaction history, and the current choices offered by the game engine. Evaluate the choices based on the following rubric:

Please assess the choices provided by the game engine based on this rubric:

[[start of rubric]]
{rubric}
[[end of rubric]]

[[start of general game instruction]]
{game}
[[end of general game instruction]]

[[start of history]]
{history}
[[end of history]]

Please assess the choices provided by the game engine:

[[start of choices]]
{choices}
[[end of choices]]

Your output should be a JSON object structured as follows:
{
    "reason": <your reasoning here>
    "score": <score from 1 to 5 based on the rubric provided>
}
You must NOT output anything else other than this JSON object.
\end{lstlisting}
\end{minipage}
\end{center}
We provide in the prompt above the game JSON~(\texttt{game}), game history up to the current round~(\texttt{history}), and three candidate choices~(\texttt{choices}). We present the following paragraphs in \texttt{rubric} for Diversity, Relevance and Understandability respectively.
\begin{center}
\begin{minipage}{0.95\textwidth}
\begin{lstlisting}[language=plaintext, frame=none, numbers=none]
Diversity: Does the set of choices provide distinct and varied options for the player?
1: The choices are nearly identical, offering no meaningful differences between them.
2: The choices have slight variations but are mostly redundant, leading to a limited sense of variety.
3: The choices exhibit some diversity but may still overlap in intent or outcome.
4: The choices are mostly distinct and provide meaningful differences that allow the player to explore different paths.
5: The choices are highly diverse, with each option offering unique and creative directions for the player.
\end{lstlisting}
\end{minipage}
\end{center}
\begin{center}
\begin{minipage}{0.95\textwidth}
\begin{lstlisting}[language=plaintext, frame=none, numbers=none]
Relevance: Are the choices appropriate and contextually aligned with the story and scene?
1: The choices are entirely irrelevant, disconnected from the scene or story, and break immersion.
2: The choices have limited relevance, with some alignment to the story but containing jarring or out-of-place elements.
3: The choices are moderately relevant, generally aligning with the story but occasionally introducing inconsistencies.
4: The choices are mostly relevant, fitting well within the context and contributing meaningfully to the story.
5: The choices are fully relevant, seamlessly integrated into the story and enhancing the narrative experience.
\end{lstlisting}
\end{minipage}
\end{center}
\begin{center}
\begin{minipage}{0.95\textwidth}
\begin{lstlisting}[language=plaintext, frame=none, numbers=none]
Understandability:  Are the choices clear, concise, and easy to understand for the player?
1: The choices are confusing, overly complex, or poorly worded, making them difficult to interpret.
2: The choices are somewhat understandable but may include ambiguous language or unnecessary complexity.
3: The choices are moderately clear, with minor ambiguities that require some interpretation.
4: The choices are clear and concise, easy to read, and free of significant ambiguity.
5: The choices are exceptionally clear and well-written, making them effortless to understand and act upon
\end{lstlisting}
\end{minipage}
\end{center}

We average these three rubric scores to obtain $act$, then normalize via $(act - 1)/4$. Trajectories are evaluated by averaging per-round scores, and we then take the mean across all trajectories:
\begin{equation}
\begin{aligned}
    \textbf{ACT}_{\text{round}} &= \frac{act-1}{4}\\
    \textbf{ACT}_{\text{traj}} &= \frac{\sum_{\text{round}} \textbf{ACT}_\text{round}}{\# rounds}\\
    \textbf{ACT} &= \frac{\sum_{\text{traj}} \textbf{ACT}_\text{traj}}{\# trajectories}
\end{aligned}.
\end{equation}

\section{Human Evaluation Details}\label{app:human_eval}
\subsection{Interface Layout}
\begin{figure}[!ht]
    \centering
    \includegraphics[width=0.95\linewidth]{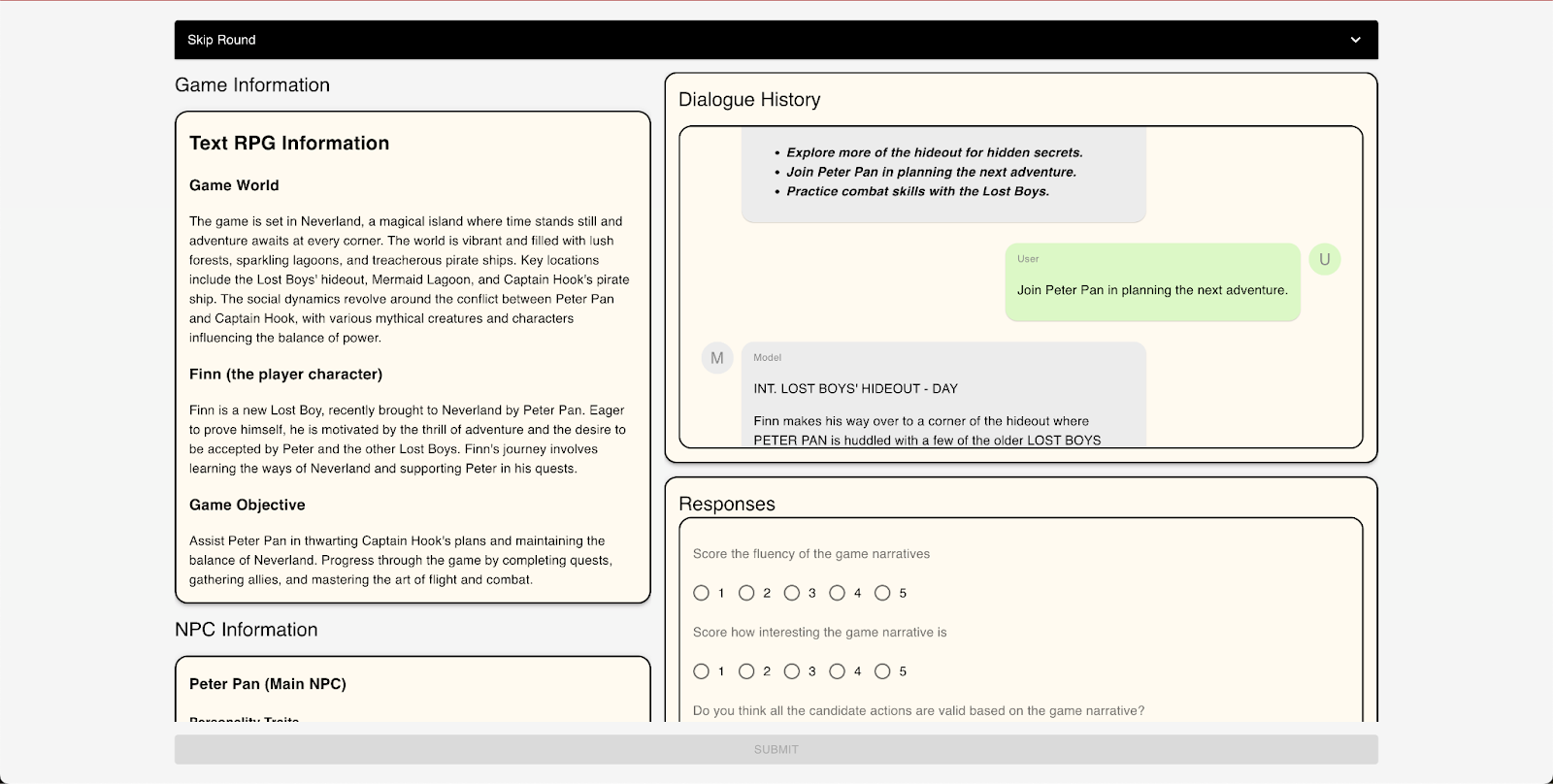}
    \caption{Screenshot of the human evaluation interface.}
    \label{fig:human-eval}
\end{figure}

Figure~\ref{fig:human-eval} shows a screenshot of our human evaluation interface. Although it is cut off due to display size, the four main components are visible: \textbf{Text RPG Information}, \textbf{NPC Information}, \textbf{Dialog History}, and \textbf{Responses}. As discussed in Appendix~\ref{app:per_eval}, we use two interfaces: one for TIPI-based personality estimation and one for direct personality-consistency evaluation. These interfaces only differ in how the \textbf{NPC Information} and \textbf{Responses} sections are presented. To help annotators remain focused when assessing a multi-round trajectory, each round in a trajectory is annotated separately by the same annotator.

\noindent\textbf{Text RPG Information:}  
Annotators see the Game World description, the player character’s name and description, and the overall game objective. This information persists throughout the trajectory.

\noindent\textbf{Dialog History:}  
We show the game trajectory up to the current round, including the model’s narration and three candidate actions (boldfaced). One of these actions, selected at random, is displayed on the right side. This component updates every round to reflect the new content.

\noindent\textbf{NPC Information:}  
For TIPI-based personality estimation, we present only the NPC’s name and facts (omitting personality traits so they can be inferred through the TIPI questions). In the direct-evaluation interface, the main NPC personality traits are included here.

\noindent\textbf{Responses:}  
This section poses natural-language questions to gather human judgments on subjective dimensions. It differs slightly between TIPI-based and direct-evaluation interfaces, as detailed below.

\subsection{Evaluation Questions}
\subsection{Evaluation Questions}
Our human evaluation asks annotators to rate various subjective aspects. Questions A--D appear every round in both the TIPI and direct-evaluation interfaces:
\begin{center}
\begin{minipage}{0.95\textwidth}
\begin{lstlisting}[language=plaintext, frame=none, numbers=none]
A. Please give a score (1-5) to indicate how interesting the game narrative is.

B. Do you think all the candidate actions are valid based on the game narrative? - 0 (no) - 1 (yes)
    
C. Are candidate choices different enough from each other, or are they essentially the same? - 0 (same) - 1 (different)
    
D. Please give a score (1-5) to measure whether the game narrative is consistent with the given facts about the main NPC? 
    - 1 has many conflicts
    - 2 has some conflicts
    - 3 neutral
    - 4 matches the description
    - 5 perfectly matches the description
\end{lstlisting}
\end{minipage}
\end{center}
These ratings inform the INT, ACT, and FAC metrics as follows:
\begin{equation}
    \begin{aligned}
        \textbf{INT}_\text{round} &= \frac{A - 1}{4},\\
        \textbf{ACT}_\text{round} &= \frac{B + C}{2},\\
        \textbf{FAC}_\text{round} &= \frac{D - 1}{4}.
    \end{aligned}
\end{equation}
Here, Question B corresponds to Relevance and Understandability in the ACT automatic evaluation, while Question C corresponds to Diversity. We average these round-level scores to obtain a trajectory-level score, then average across all trajectories.

\paragraph{Personality Consistency Questions (E1 and E2).}
We measure PER using two different question sets:
\begin{itemize}
    \item \textbf{E1: TIPI Estimation.}  
    Shown only once per trajectory (at the final round of the TIPI interface), requiring annotators to assess the entire trajectory.  
    \item \textbf{E2: Direct Evaluation.}  
    Appears at every round in the direct-evaluation interface.
\end{itemize}

Both methods yield PER scores analogous to the automatic evaluations in Appendix~\ref{app:per_eval}.

\begin{center}
\begin{minipage}{0.95\textwidth}
\begin{lstlisting}[language=plaintext, frame=none, numbers=none]
E1. Here are a number of personality traits that may or may not apply to the character. Please write a number to each statement to indicate the extent to which you agree or disagree with that statement, based ONLY on the game narratives. You should rate the extent to which the pair of traits applies to the character, even if one characteristic applies more strongly than the other. Use a score range of 1-7:
    - 1: Disagree strongly
    - 2: Disagree moderately
    - 3: Disagree a little
    - 4: Neither agree nor disagree
    - 5: Agree a little
    - 6: Agree moderately
    - 7: Agree strongly

    I see the main NPC as
    A. Extraverted, enthusiastic.
    B. Critical, quarrelsome.
    C. Dependable, self-disciplined.
    D. Anxious, easily upset.
    E. Open to new experiences, complex.
    F. Reserved, quiet.
    G. Sympathetic, warm.
    H. Disorganized, careless.
    I. Calm, emotionally stable.
    J. Conventional, uncreative

E2. Please give a score (1-5) to measure whether the game narrative is consistent with the given facts about the main NPC?
    - 1 has many conflicts
    - 2 has some conflicts
    - 3 neutral
    - 4 matches the description
    - 5 perfectly matches the description
\end{lstlisting}
\end{minipage}
\end{center}
\subsection{Annotation Setup}
We recruited 15 human annotators. Each trajectory is annotated at the round level, resulting in two annotations per interface type and therefore four total annotations per trajectory. We ensure that each annotator encounters any given trajectory only once, regardless of interface type. Consequently, each trajectory ends up with four sets of INT, ACT, and FAC scores, and two sets of PER and \textbf{PER}$^d$ scores. We take the mean over all trials to produce the final reported values. For inter-annotator agreement (Table~\ref{tab:gs_corr}), we randomly divide the collected annotations into two groups and compare their scores.

\subsection{PER vs. PER$^d$ Evaluation Results}
\begin{table*}[!ht]
\centering
\begin{tabular}{lrrrrrr}
\toprule
\multirow{2}{*}{Model} & \multicolumn{2}{c}{PER (Subset)} & \multicolumn{2}{c}{PER$^d$ (Subset)} & \multicolumn{1}{c}{PER (Full)} & \multicolumn{1}{c}{PER$^d$ (Full)}\\ 
& \multicolumn{1}{c}{Auto} & \multicolumn{1}{c}{Human} & \multicolumn{1}{c}{Auto} & \multicolumn{1}{c}{Human} & \multicolumn{1}{c}{Auto} & \multicolumn{1}{c}{Auto}\\
\midrule
Claude 3.5 Sonnet & 0.729 & 0.648 & 0.768 & \textbf{0.832} & 0.589 & 0.738 \\
Deepseek V3 & 0.742 & 0.645 & 0.750 & \underline{0.826} & 0.583 & \textbf{0.778}\\
Gemini 1.5 Pro & 0.740 & 0.648 & \textbf{0.800} & 0.769 & \underline{0.596} & \underline{0.777}\\
Gemini 2.0 Flash Exp & 0.737& \underline{0.651} & 0.707 & 0.769 &\textbf{0.598} & 0.750\\
GPT 4o & 0.711 & \textbf{0.667} & 0.780 & 0.724 & 0.585 & 0.768 \\
GPT 4o mini & \textbf{0.753}  & 0.648 & \underline{0.788} & 0.735 & 0.588 & 0.763\\
Llama 3.1 70B & \underline{0.744}  & 0.627 & 0.768 & 0.752 & 0.586 & 0.765\\
Llama 3.3 70B & 0.739 & 0.640 & 0.739 & 0.755 & 0.585 &0.774\\
\bottomrule
\end{tabular}
\caption{PER and PER$^d$ results from automatic and human evaluation on a subset of 20 games, and automatic evaluation on the full set of games.}
\label{tab:per_evaluation_results}
\end{table*}

\begin{table}[!ht]
    \centering
    \begin{tabular}{cc|rrr}
    \toprule
    \multicolumn{2}{c|}{Comparisons} & Pearson & Kendall & MAD \\
    \midrule
    \multirow{1}{*}{Auto-Auto Agreement} & PER Auto - PER$^d$ Auto & 0.013 & 0.109 & 0.037 \\
    \midrule
    \multirow{3}{*}{Auto-Human Agreement} & PER Auto - PER Human     & -0.691 & -0.429 & 0.090\\
    & PER$^d$ Auto - PER$^d$ Human &-0.297 & -0.255 & 0.047\\
    \midrule
    \multirow{3}{*}{Human-Human Agreement} &PER Human - PER Human & -0.310 & -0.286 & 0.023\\
    &PER$^d$ Human - PER$^d$ Human & 0.649 & 0.143 & 0.035\\
    &PER Human - PER$^d$ Human &-0.175&-0.143 & 0.124\\
    \bottomrule
    \end{tabular}
    \caption{Agreement analysis for PER and PER$^d$ scores. We present Pearson correlation coefficient (Pearson), Kendall rank correlation coefficient (Kendall), and Mean Absolute Difference (MAD)}
    \label{tab:per_agreement}
\end{table}

In our main article, we adopt the PER score for evaluating NPC personality consistency. Here, we further analyze both PER and PER$^d$ scores from automatic and human evaluations on a subset of 20 games in Table~\ref{tab:per_evaluation_results}, with additionally automatic evaluation results on the full dataset. We also report agreement metrics in Table~\ref{tab:per_agreement} Our analysis reveals several key observations:

\begin{enumerate}
    \item \textbf{PER$^d$ tends to be higher than PER in both automatic and human evaluations.} Across models, we observe that PER$^d$ scores are consistently higher than PER scores, indicating that direct evaluation of personality consistency is generally more lenient than the TIPI-based method. This trend holds for both automatic and human evaluators.

    \item \textbf{LLMs achieve similar PER scores across the dataset.} The automatic PER and PER$^d$ scores on the full set of games show little variation across models, with all models achieving scores around 0.58–0.60 for PER and around 0.74–0.78 for PER$^d$. This suggests that models perform comparably in terms of maintaining personality consistency in text-based role-playing.

    \item \textbf{Human evaluators rate PER$^d$ higher than PER, but with noticeable variation.} While automatic evaluations show a clear gap between PER and PER$^d$, human annotations exhibit a similar pattern but with greater variability. Notably, human evaluators assign significantly higher PER$^d$ scores to some models, such as Claude 3.5 Sonnet and DeepSeek V3, compared to their automatic scores.

    \item \textbf{Human and automatic PER scores exhibit poor correlation.} Table~\ref{tab:per_agreement} shows that the Pearson correlation between PER Auto and PER Human is negative (-0.691), with Kendall correlation also negative (-0.429). This suggests a fundamental mismatch between how LLM-based and human evaluators assess personality consistency through TIPI.

    \item \textbf{Better human agreement for PER$^d$, but still unstable.} While inter-human correlation for PER is negative (-0.310 Pearson, -0.286 Kendall), PER$^d$ exhibits a stronger but still weak agreement (0.649 Pearson). This suggests that directly rating personality alignment may be more intuitive for human evaluators than using TIPI scores but remains somewhat unstable. However, there is still concern over whether human annotators are capable of accurately understanding Big Five traits in the direct evaluation scenario.

    \item \textbf{Low agreement between PER and PER$^d$.} The Pearson correlation between PER and PER$^d$ scores (both automatic and human) is low (0.013 for Auto-Auto and -0.175 for Human-Human), indicating that these two evaluation methods capture different aspects of personality consistency. While PER$^d$ measures direct alignment with given traits, PER (TIPI) estimates personality traits implicitly, which may introduce more variance in judgments.
\end{enumerate}

\paragraph{Justification for Choosing TIPI (PER) in the Main Article.}  
We adopt TIPI-based personality consistency (\textbf{PER}) rather than direct evaluation (\textbf{PER}$^d$) in the main study for several reasons. First, TIPI does not require evaluators to have prior knowledge of the Big Five personality traits, making it a structured and interpretable method for assessing personality consistency. Additionally, the high variance in PER$^d$ human scores (as seen in Table~\ref{tab:per_agreement}) suggests that direct personality evaluation is more susceptible to subjective biases. The negative correlation between automatic and human PER scores further emphasizes the challenge of aligning LLM-based and human-based assessments, reinforcing the need for a more systematic approach like TIPI.

Overall, these results highlight the complexity of evaluating personality consistency, where different evaluation paradigms yield divergent results. The instability in human-human agreement for both PER and PER$^d$ suggests that subjective evaluation remains a challenging aspect of LLM benchmarking, warranting further research into more reliable personality evaluation methodologies.


\end{document}